\pdfoutput=1

\documentclass[11pt]{article}

\usepackage[final]{acl}

\usepackage{times}
\usepackage{latexsym}
\usepackage{colortbl}
\usepackage{multirow}
\usepackage{booktabs}
\usepackage[many]{tcolorbox}

\usepackage[T1]{fontenc}

\usepackage[utf8]{inputenc}

\usepackage{microtype}
\usepackage{graphicx}

\title{ConQuer: A Framework for \underline{Con}cept-Based \underline{Qu}iz Gen\underline{er}ation}

\author{
  Yicheng Fu$^1$, Zikui Wang$^1$, Liuxin Yang$^1$, Meiqing Huo$^1$, and Zhongdongming Dai$^2$ \\
  $^1$Stanford University, CA, USA \\
  $^2$University of California San Diego, CA, USA \\
  \{easonfu, zikuiw, lyang822, meiqing7\}@stanford.edu$^1$ \\
  z1dai@ucsd.edu$^2$
}
\begin{document}
\maketitle

\begin{abstract}
Quizzes play a crucial role in education by reinforcing students' understanding of key concepts and encouraging self-directed exploration. However, compiling high-quality quizzes can be challenging and require deep expertise and insight into specific subject matter. Although LLMs have greatly enhanced the efficiency of quiz generation, concerns remain regarding the quality of these AI-generated quizzes and their educational impact on students. To address these issues, we introduce \textbf{ConQuer}, a concept-based quiz generation framework that leverages external knowledge sources. We employ comprehensive evaluation dimensions to assess the quality of the generated quizzes, using LLMs as judges. Our experiment results demonstrate a 4.8\% improvement in evaluation scores and a 77.52\% win rate in pairwise comparisons against baseline quiz sets. Ablation studies further underscore the effectiveness of each component in our framework. Code available at \url{https://github.com/sofyc/ConQuer}.


\end{abstract}

\section{Introduction}

Quizzes are a widely used tool in modern education, serving as a means to test students' understanding of material and providing opportunities for reflection~\cite{cheong2013quick, evans2021effectiveness}. Well-designed quizzes can enhance active learning, provide valuable feedback, and stimulate curiosity ~\cite{malandrino2014quiz, mukaromah2019digital}. However, the process of creating quizzes is often labor intensive, requiring subject matter expertise, careful consideration of key concepts, and understanding of students' knowledge levels ~\cite{gorin2006test}. This challenge becomes even more pronounced in fields where content is updated frequently or where educators need to generate quizzes on a scale.

In recent years, the emergence of Large Language Models has provided a promising solution to these challenges. LLMs can quickly generate quizzes that cover a wide range of topics. ~\citet{elkins2023useful} demonstrated that the LLM-generated quizzes are promising for widespread use in the classroom. Although this approach offers significant efficiency gains, it also raises concerns about the quality and relevance of the generated quizzes \cite{lodovico2024comparison}. Specifically, there are questions about whether the quizzes accurately reflect key concepts in a given domain and whether they are grounded in reliable sources of knowledge~\cite{zhang2023siren}.

To address these concerns, we propose a concept-based quiz generation method grounded in external knowledge corpora, such as Wikipedia and ConceptNet ~\cite{conceptnet}. Using concepts instead of keywords to search for relevant information enables the capture of knowledge points that may not be explicitly mentioned in students' questions. By anchoring quiz generation in well-established knowledge bases, our approach ensures that quizzes are not only relevant but also comprehensive, covering critical concepts that learners must grasp. 



We employ comprehensive evaluation dimensions to assess various aspects of quiz quality. Our concept-based approach achieves a 4.8\% improvement in evaluation scores compared to traditional LLM-generated quizzes. In pairwise evaluations, our method consistently outperforms other alternatives,with 77.52\% of evaluations favoring our method over LLM-generated quizzes. Additionally, ablation studies reveal the critical contributions of the concept extraction module, knowledge source, and summary module in enhancing the overall effectiveness of our framework.

\begin{figure*}
    \centering
    \includegraphics[width=\linewidth]{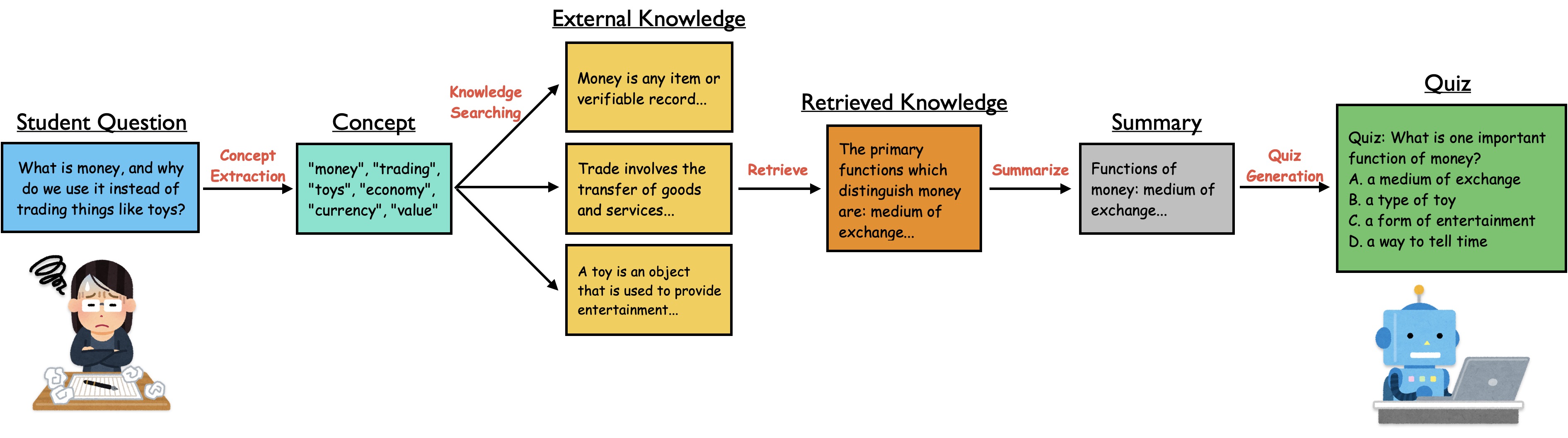}
    \caption{The ConQuer Framework. First, key concepts are extracted from student questions, followed by retrieving relevant information from external knowledge sources based on semantic similarity. Finally, the main topics are summarized to generate personalized quizzes.}
    \label{fig:3}
\end{figure*}

In summary, our key contributions are as follows:\begin{itemize} 
    \item We present \textbf{ConQuer}, a novel concept-based quiz generation framework that significantly improves the quality of LLM-generated quizzes. A diagram of our framework is shown in Figure~\ref{fig:3}.

    \item We conduct a detailed ablation study with qualitative analysis, revealing that each component of our framework plays a crucial role in improving the quality of quiz generation. 

     \item We release our student question dataset and quiz generation pipeline code as open-source resources to facilitate future research.
\end{itemize}

\section{Related Work}

\paragraph{Retrieval-Augmented Generation} While LLMs have demonstrated strong performance in various understanding and reasoning tasks, their ability to generate reliable and factually accurate text remains a challenge, particularly in knowledge-intensive tasks ~\cite{kandpal2023large}. This often leads to hallucinations, where models produce incorrect or fabricated information ~\cite{zhang2023siren}. Retrieval-Augmented Generation (RAG) ~\cite{lewis2020retrieval} has been proposed to address this issue by integrating LLMs with retrieval mechanisms, allowing models to refer to external databases and improve the factuality and credibility of their outputs. RAG has shown promising results in QA tasks such as SQuAD~\cite{rajpurkar2018know} and HotpotQA~\cite{yang2018hotpotqa}, as well as in personal planning applications~\cite{fu2024camphor}, where external knowledge is essential for generating accurate and contextually relevant responses.

\paragraph{Quiz Generation} Quizzes are Widely recognized as an effective tool to promote active learning and improve knowledge retention ~\cite{evans2021effectiveness, mukaromah2019digital}. Recent studies have explored how large language models (LLMs) can be used to improve the quality of generated quiz content. For instance, ~\citet{vu2024chatgpt} investigates interactive prompting strategies for designing question banks, while ~\citet{hasan2024automatic} combines LLMs with structured resources to enhance factual accuracy and contextual relevance in quiz generation. Additionally, ~\citet{gabajiwala2022quiz} explores keyword extraction to generate better quizzes. ~\citet{biancini2024multiple} proposes to generate quizzes by Injecting external knowledge into LLM prompts. These approaches typically rely on pre-identified topic and keyword-based techniques. In contrast, our ConQuer framework tackles scenarios where Students may lack awareness of the concepts they need to learn, requiring a focus on deeper concept identification rather than surface-level keyword-based methods.

\section{Task}

Previous studies have explored quiz generation based on predefined topics \cite{song2016domain, vu2024chatgpt}. However, Such topic-centered approaches often fails to capture the complexities of real-world educational settings. In practice, students Frequently ask vague or incomplete question, sometimes without fully grasping the underlying concepts they are struggling with ~\cite{commeyras1995can}. Research in education has shown that students' questions can reflect their thought processes and serve as a valuable resource to enhance learning \cite{cuccio2000enhancement, chin2008students}. Inspired by this, our approach shifts from relying on predefined topics for LLM-based quiz generation. Instead, we focus on generating questions that mirror the types of inquiries students might pose to instructors, capturing their authentic learning challenges. The task then becomes generating quizzes that effectively support students with limited information about their current knowledge level.

\begin{figure*}[ht]
    \centering
    \begin{minipage}{0.5\textwidth}
        \centering
        \includegraphics[width=\linewidth]{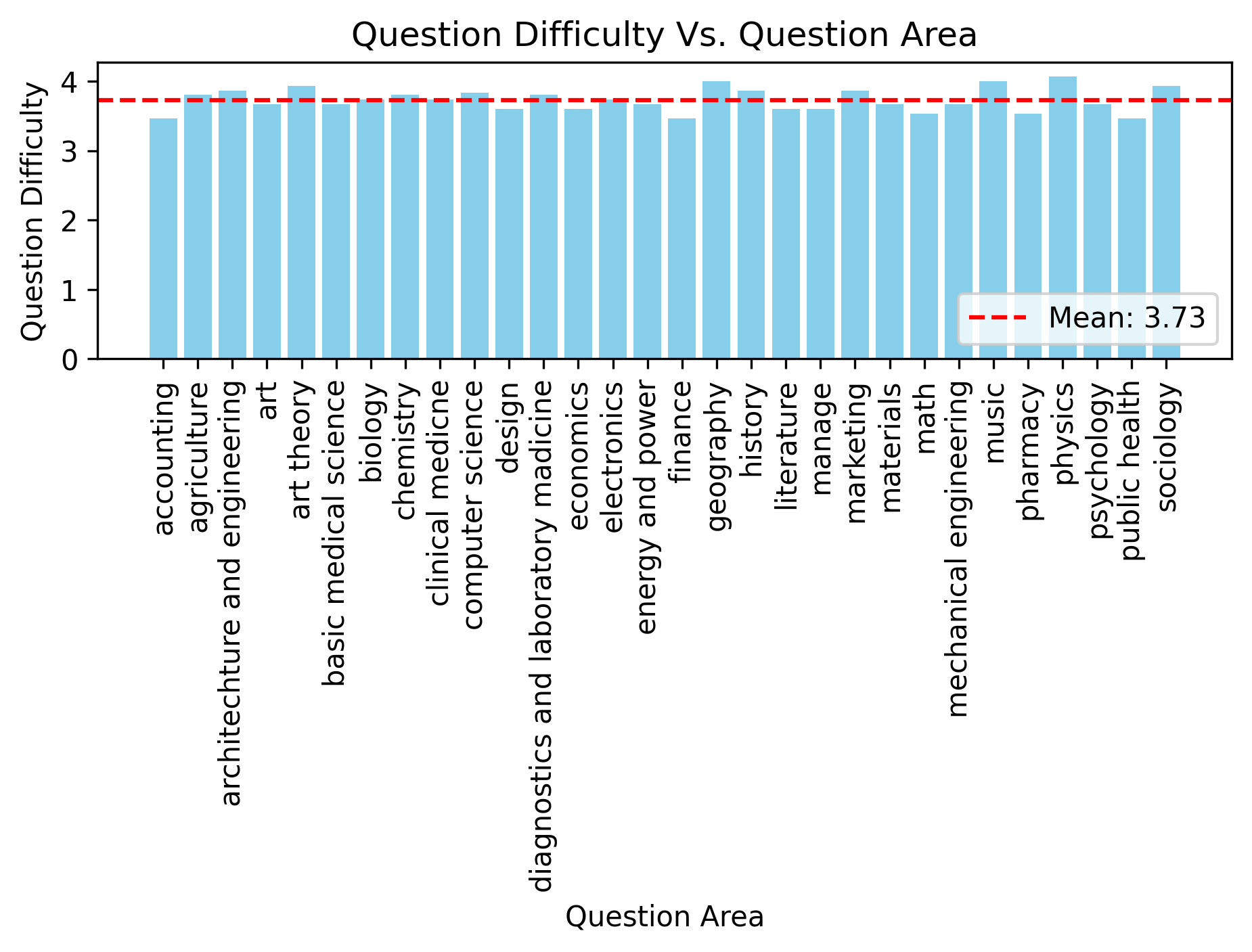}
        \caption{Student Question Difficulty Vs. Area}
        \label{fig:1}
    \end{minipage}%
    \begin{minipage}{0.5\textwidth}
        \centering
        \includegraphics[width=\linewidth]{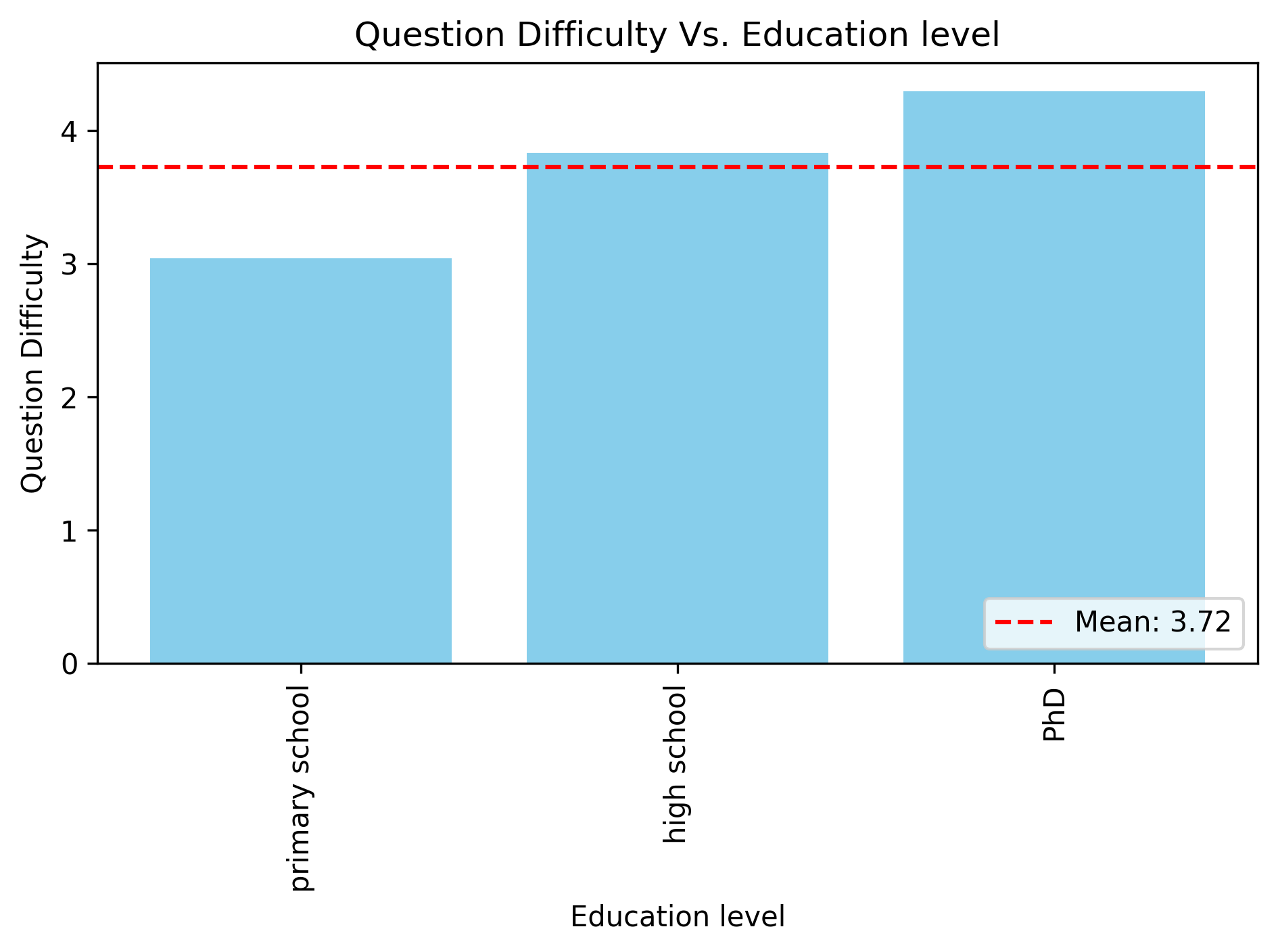}
        \caption{Student Question Difficulty Vs. Education Level}
        \label{fig:2}
    \end{minipage}
\end{figure*}

To enhance the diversity of student questions and broaden the framework's applicability across a wide audience, we selected 30 subject areas from the MMLU dataset~\cite{mmlu} and considered three educational levels: primary school, high school, and PhD. For each subject and educational level, we tasked GPT-4o~\cite{hurst2024gpt} with generating five representative questions that students would typically ask. This approach yields a dataset of 450 questions, which we compiled into a comprehensive question set for experiment. The quiz generation task involves generating three quizzes for each student question, where each quiz consists of one question, one correct answer, and three incorrect options, with the correct answer always positioned as option A. We believe that a single quiz may only provide a limited perspective on the topic, and a set of quizzes offers a more comprehensive approach, thereby enhancing students' overall understanding of the subject matter.

To verify that the difficulty of the student questions varies appropriately across different education levels, we tasked the LLM with assessing the reasoning difficulty and knowledge depth required to answer each student question, assigning a score on a scale of 1 to 5. The results are presented in Figures~\ref{fig:1} and ~\ref{fig:2}. As anticipated, the difficulty remains consistent within subject areas but increases progressively with the educational level, aligning with our goal for the dataset. Example student questions can be found in the Appendix ~\ref{apd:box}.

\section{Framework}

The proposed framework ConQuer operates as follows. The system receives three inputs: the student’s question, their educational level, and the subject area. We first use an LLM to extract key concepts from the question. For example, given the question, "What happens to a plant when it doesn't get enough sunlight or water?", we identify several potentially relevant key concepts such as "plant", "sunlight", "water", "photosynthesis", "growth", "stress", "environment".

After extracting the relevant concepts, we retrieve relevant information from a knowledge source based on these concepts. In this work, we primarily use Wikipedia, which provides a wealth of information on a wide range of topics. To locate the most relevant content, we utilize Sentence-BERT ~\cite{reimers2019sentence} to compute cosine similarity scores. This allows the system to pinpoint the most contextually appropriate sections of text. Subsequently, an LLM-based summarization module condenses the retrieved information into its key points. These summarized details are then passed to the quiz generator, which creates tailored quizzes based on the content.

\section{Experiments}

\subsection{Evaluation}

To evaluate the quality of the generated quizzes, we propose 5 evaluation dimensions: 
\begin{itemize} 
    \item \textbf{Educational Value}: Whether the quizzes enhance learning and help students acquire new knowledge. 
    \item \textbf{Diversity}: Whether the quizzes cover a broad range of important topics and concepts. 
    \item \textbf{Area Relevance}: How well the quizzes align with the student’s query and the specific subject area they are trying to learn. \item \textbf{Difficulty Appropriateness}: Whether the quiz difficulty matches the student's education and knowledge level. 
    \item \textbf{Comprehensiveness}: Whether the quizzes cover the topic’s key concepts thoroughly. 
\end{itemize}

We leverage LLM-as-a-judge for evaluation, with GPT-4o ~\cite{hurst2024gpt} serving as the judge model in all our evaluations. The model is instructed to assign a score on a scale of 1 to 5, with detailed prompts provided in Appendix~\ref{apd:pro}. To further compare quiz quality, we perform pairwise comparisons, prompting the judge model to select the better quiz set based on each of the five criteria outlined above. To mitigate any potential ordering bias, pairwise comparisons are conducted in both orders, and the average win rate is computed.

\subsection{Experiment Setup}

We use GPT-4o-mini~\cite{hurst2024gpt} and Gemini-2.0-flash~\cite{team2023gemini} as LLMs to complete the task. For information retrieval, we employ the text-embedding-3-large model for embeddings, with a chunk size of 128, a chunk overlap of 50, and retrieve the top 3 results. Detailed prompts are provided in the Appendix~\ref{apd:pro}.

\subsection{Ablation Study}

To evaluate the contribution of each component in our framework, we conduct three ablation studies.

\begin{itemize}
    \item \textbf{Concept Extraction Module:} We remove the concept extraction module and rely solely on the pure words from the sentence after removing stop words and punctuation to search for relevant information in Wikipedia.

    \item \textbf{Knowledge Source:} Instead of retrieving information from Wikipedia, we rely on ConceptNet~\cite{conceptnet} to gather related concepts and their relational descriptions in sentence format. Unlike Wikipedia, which provides detailed introductions to each term, ConceptNet only includes simple relational descriptions between words like "Find [[a money]] in [[a bank]]".

    \item \textbf{Summarization Module:} We remove the summarization module and directly feed all the information retrieved from Wikipedia into the quiz generator without any further processing.
\end{itemize}

\begin{table*}[ht!]
\centering
\resizebox{0.9\textwidth}{!}{
\begin{tabular}{lccccccc}

\toprule
\textbf{Source} & \textbf{EV} & \textbf{Diversity} & \textbf{AR} & \textbf{DA} & \textbf{Comprehensiveness} & \textbf{Avg} & \textbf{$\Delta$}\\
\midrule
\textbf{ConQuer} & \textbf{83.22} & 52.04 & \textbf{97.18} & \textbf{84.70} & \textbf{61.34} & \textbf{75.70} & --- \\
\midrule
\textbf{\textit{- Concept Extraction}} & 80.31 & 52.13 & 93.24 & 83.36 & 59.42 & 73.69 & -2.66\% \\
\textbf{\textit{ConceptNet}} & 80.36 & \textbf{53.06} & 93.60 & 83.40 & 59.91 & 74.07 & -1.32\% \\
\textbf{\textit{- Summary}} & 77.49 & 52.39 & 88.28 & 81.88 & 57.94 & 71.60 & -5.42\% \\
\bottomrule
\end{tabular}
}
\caption{Ablation Study Results. EV, AR, and DA stand for Educational Value, Area Relevance, and Difficulty Appropriateness, respectively.}
\label{tab:ablation}
\end{table*}

\section{Results}

\begin{figure}[ht]
    \centering
    \includegraphics[width=\linewidth]{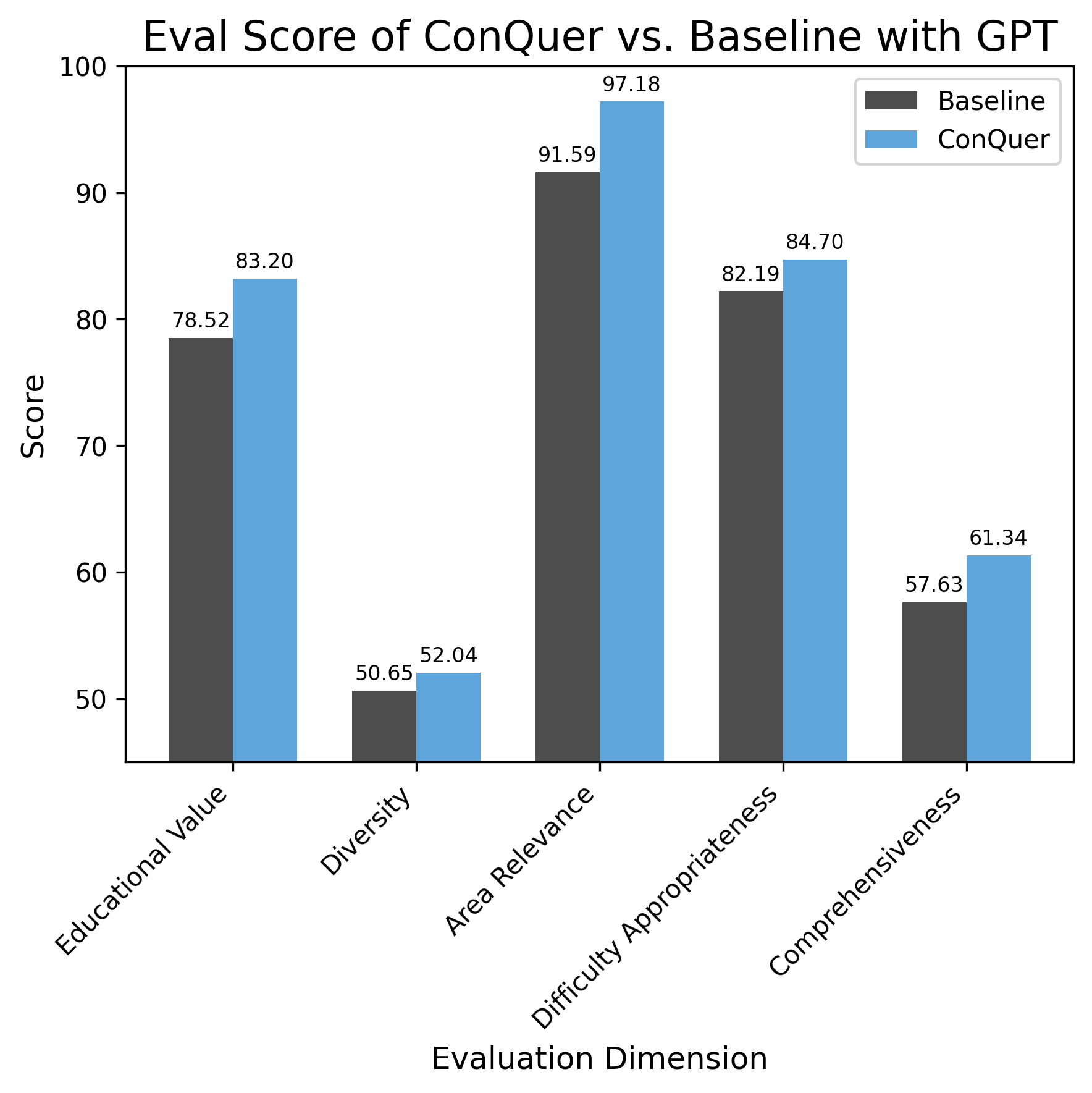}
    \caption{Evaluation score comparison between the baseline and ConQuer with GPT-4o-mini. The evaluation score has been normalized to a scale of 100.}
    \label{fig:4}
\end{figure}

\begin{figure}[ht]
    \centering
    \includegraphics[width=\linewidth]{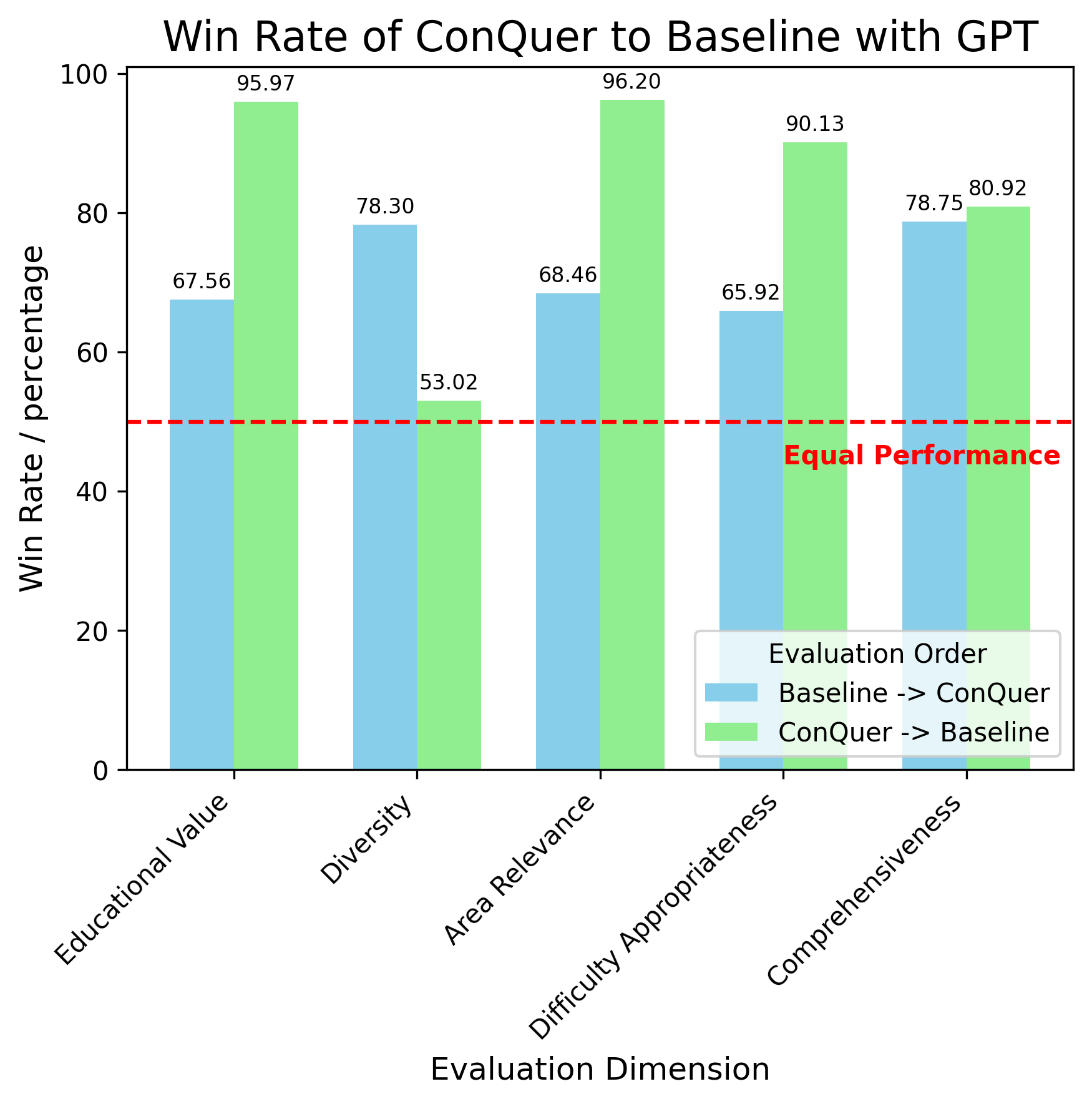}
    \caption{Win rate from pairwise comparison between the baseline and ConQuer with GPT-4o-mini}
    \label{fig:5}
\end{figure}

We compare the performance of our ConQuer framework against a baseline, where the quiz is generated directly from the student's question without utilizing any external materials or concepts. The evaluation score for quizzes generated by GPT-4o-mini is shown in Figure~\ref{fig:4}, and the win rate of ConQuer in pairwise comparison is presented in Figure~\ref{fig:5}. For clarity, the evaluation score has been scaled to 100, and the win rate is expressed as a percentage. Additional results for Gemini-2.0-flash can be found in the Appendix ~\ref{apd:exp}.

Our results, based on both evaluation score and pairwise comparison win rates, demonstrate that ConQuer consistently outperforms the baseline across all five evaluation dimensions. Although the average score improvement across these five dimensions is only 4.8\%, ConQuer achieves a significant win rate of 77.52\% in the pairwise comparison. We hypothesize that both ConQuer and the baseline produce quizzes that appear well-constructed when considered in isolation, leading to high evaluation scores for both. However, when evaluated together, ConQuer significantly outperforms the baseline because its quizzes are grounded in high-quality knowledge sources that are closely aligned with key concepts. We have observed that the LLM judge tends to prefer the second candidate, except in the diversity dimension. This preference may be attributed to the fact that the second candidate is closer to answer tokens, prompting the model to allocate more attention weights to it. In the diversity dimension, however, the model identifies more repeated content in the second candidate, as it has already seen all the quizzes from the first candidate. Correlation analysis of the five evaluation dimensions can be found in Appendix ~\ref{apd:cor}.

\subsection{Results of Ablation Studies}

The results of ablation study are in Table ~\ref{tab:ablation}. We observe that removing any of the three components leads to a decrease in performance, except in the diversity dimension. Although the performance drop is minimal, as noted in the previous analysis, it may represent a significant reduction in quality when compared to the original quiz. The diversity score remains largely unaffected, likely because the task only requires generating three quizzes, making it relatively easy to ensure variety. A qualitative analysis of each ablation experiment is provided in the Appendix ~\ref{apd:qua}.

Removing the concept extraction module leads to the loss of important concepts that may not be explicitly mentioned in the sentence. For example, in the student question, "What happens to a plant when it doesn't get enough sunlight or water?", the key concept "Photosynthesis" is missing, resulting in the omission of vital information.

Using ConceptNet as the knowledge source reduces the richness and quality of the retrieved information, although this is not as apparent in the three-quiz scenario since the retrieved information is still sufficient to generate distinct quizzes.

Removing the summarization module causes the most significant drop in scores. This likely happens because the model is overwhelmed by the excessive information and struggles to focus on the key elements.

\section{Conclusion}
We introduced ConQuer, a concept-based framework for generating conceptually grounded and educationally effective quizzes. By prioritizing key concepts over surface-level keywords, ConQuer ensures alignment with essential learning objectives. Our evaluations show a 4.8\% improvement in quiz quality and a 77.52\% win rate in pairwise comparisons, highlighting the superiority of our approach. Ablation studies emphasize the importance of each component in driving these improvements.

ConQuer offers a scalable, accurate, and pedagogically valuable tool for quiz generation across diverse educational contexts. Future work could extend knowledge sources, refine quiz generation for adaptive difficulty, and personalize learning path. ConQuer represents a step forward in automating quiz creation while ensuring the accuracy and relevance that are critical to effective learning.

\clearpage

\section*{Limitations}
While ConQuer demonstrates significant improvements over the baseline in several aspects, there are key limitations to our evaluation. One notable limitation is that it is evaluated for generating only three multiple-choice quizzes, limiting its generalizability to other quiz formats or larger-scale quiz settings. 

Another limitation is that our evaluation relies solely on LLMs for assessing quiz quality, without human input, which may undermine its validity by omitting human values and preferences. Additionally, the lack of feedback assessment limits the practical usefulness of the quizzes. Future research should explore the impact of personalized quiz generation based on student profiles, such as learning history and preferences.

In interactive learning environments, students often expect quizzes to be generated rapidly; however, the inherent latency of LLMs can hinder this expectation. Addressing this challenge may require integrating supplementary LLM serving systems with adaptive computing strategies, as proposed in~\cite{fu2024efficiently}.

Finally, while concept extraction plays a crucial role, it is not without its flaws. Critical concepts may be overlooked or misinterpreted, particularly when questions are ambiguous or contain implicit ideas, potentially compromising quiz quality and relevance.

\bibliography{anthology,custom}

\clearpage
\appendix

\section{Example student question}
\label{apd:box}

Example questions of biology area can be found below. The question difficulty across the three educational levels shows a clear progression. For primary school, questions focus on basic understanding of natural phenomena and straightforward cause-and-effect relationships, requiring minimal specialized knowledge. At the high school level, questions become more complex, involving scientific methods, deeper concepts like genetic variation and the impacts of human activities, and a higher demand for critical thinking. For PhD-level questions, the focus shifts to advanced research topics, such as methodologies in studying microbiomes and the ethical implications of genetic manipulation.

\begin{tcolorbox}[colframe=green!50!black, colback=gray!10, title=Example Student Questions, breakable]
\textbf{Primary School:}
\begin{itemize}
    \item What are the ways plants and animals adapt to their environments to survive?
    \item How do some animals use camouflage to protect themselves from predators?
    \item What happens to a plant when it doesn't get enough sunlight or water?
    \item Why do some animals migrate long distances, and how do they find their way?
    \item How do different animal habitats, like forests and deserts, affect the types of species that live there?
\end{itemize}

\textbf{High School:}
\begin{itemize}
    \item What are the various methods scientists use to study ecosystems, and what challenges do they face in collecting data?
    \item How do genetic variations within a population contribute to natural selection and evolution?
    \item What role do enzymes play in biochemical reactions, and how can temperature and pH affect their activity?
    \item In what ways do human activities impact biodiversity, and what strategies can be employed to mitigate these effects?
    \item How do different types of symbiotic relationships (like mutualism and parasitism) influence ecological balance?
\end{itemize}

\textbf{PhD:}
\begin{itemize}
    \item What are the current methodologies used in studying the microbiome's influence on human health, and how do they differ in their approaches?
    \item How does epigenetic modification play a role in the adaptation of organisms to their environments over generations?
    \item What are the key differences in the mechanisms of action between CRISPR technologies and traditional gene editing techniques?
    \item In studying evolutionary biology, how do we measure and interpret the rate of speciation in various ecosystems?
    \item What ethical considerations arise in the manipulation of genetic material in research, particularly regarding biodiversity conservation?
\end{itemize}
                
\end{tcolorbox}

\section{Prompts}
\label{apd:pro}

Here is the prompts we use as baseline method
\begin{tcolorbox}[colframe=black, colback=gray!10, title=Baseline Prompt, breakable]

You are a quiz generator. The students are currently studying \{area\} at the \{level\} level and have asked a question. Your task is to create 3 quizzes that help the student better understand the question. The quiz should consist of one question, one correct answer, and three incorrect options. The correct answer must always be placed in option A. \\

Example: \\

Student Question: Where is Beijing located? \\
$[$Quiz$]$ \\
Quiz: What is the capital city of China? \\
A. Beijing \\
B. Chengdu \\
C. Shanghai \\
D. Hangzhou \\

$[$Quiz$]$ \\
Quiz: What continent is Beijing located? \\
A. Asia \\
B. Europe \\
C. Africa \\
D. North America \\

Now, please generate 3 quizzes following the format, each quiz should follow the sign of [Quiz]: \\
Student Question: \{question\}
\end{tcolorbox}

Here is the prompt we use with WikiPedia knowledge:
\begin{tcolorbox}[colframe=black, colback=gray!10, title=ConQuer Prompt, breakable]

You are a quiz generator. The students are currently studying \{area\} at the \{level\} level and have asked a question. Your task is to create 3 quizzes that helps the student better understand the question. You have access to summarized reference information from Wikipedia. The quizzes should accurately reflect reference information, and the correct answer must be well-supported by reference information. The quiz should consist of one question, one correct answer, and three incorrect options. The correct answer must always be placed in option A. \\

Example: \\

Student Question: Where is Beijing located? \\
$[$Quiz$]$ \\
Quiz: What is the capital city of China? \\
A. Beijing \\
B. Chengdu \\
C. Shanghai \\
D. Hangzhou \\

$[$Quiz$]$ \\
Quiz: What continent is Beijing located? \\
A. Asia \\
B. Europe \\
C. Africa \\
D. North America \\

Now, please generate 3 quizzes following the format, each quiz should follow thw sign of $[$Quiz$]$: \\

Reference Wikipedia Information: \\
\{summary\}

Student Question: \{question\}
\end{tcolorbox}

Here is the prompt we use to evaluate the overall quality of quiz set:
\begin{tcolorbox}[colframe=black, colback=gray!10, title=Prompt for Quiz Quality Evaluation, breakable]

A student studying \{area\} at the \{level\} level is asking a question: "\{question\}". Based on the following quiz set related to the question, I need you to evaluate the educational quality of the quiz set.  For each of the following criteria, assign a score from 1 to 5 for the entire quiz set: \\

1. Educational Value: Do you think these quizzes are educational? Will students learn more by taking these quizzes? \\
    - 1: Not educational at all, no learning value. \\
    - 2: Minimally educational, little learning value. \\
    - 3: Moderately educational, some learning value. \\
    - 4: Very educational, strong learning value. \\
    - 5: Highly educational, great learning value. \\

2. Diversity: Do you think these quizzes are diverse? Are the quizzes covering a broad range of topics, or do they all focus on the same concept? \\
    - 1: Very repetitive, covers a narrow area. \\
    - 2: Some diversity, but mostly focuses on one concept. \\
    - 3: Fairly diverse, covers a few different topics. \\
    - 4: Quite diverse, covers multiple relevant topics. \\
    - 5: Extremely diverse, covers a broad range of topics. \\

3. Area Relevance: Are these quizzes relevant to the student's question and the concepts they're trying to learn? Are the quizzes tailored to the subject area being studied? \\
    - 1: Not relevant to the question or subject at all. \\
    - 2: Minimally relevant, some connection to the question/subject. \\
    - 3: Moderately relevant, fairly aligned with the question/subject. \\
    - 4: Highly relevant, strongly aligned with the question/subject. \\
    - 5: Perfectly relevant, directly tied to the question/subject. \\

4. Difficulty Appropriateness: Do you think these quizzes match the student's current education level? Would these quizzes be too easy or too difficult for a student at this level? \\
    - 1: Too easy or too difficult, not appropriate for the level. \\
    - 2: Slightly mismatched, quizzes may be too easy or too hard. \\
    - 3: Moderately appropriate, quizzes are somewhat aligned with the level. \\
    - 4: Mostly appropriate, quizzes are well-suited for the level. \\
    - 5: Perfectly suited to the student's education level. \\

5. Comprehensiveness: Do these quizzes cover the depth and breadth of the topic? Are they thorough in addressing key concepts and details? \\
    - 1: Very superficial, only scratches the surface of the topic. \\
    - 2: Somewhat incomplete, misses important aspects. \\
    - 3: Moderately comprehensive, covers the basics but lacks depth. \\
    - 4: Quite comprehensive, addresses most key aspects with reasonable depth. \\
    - 5: Highly comprehensive, thoroughly covers the topic in great depth and detail. \\

Here is the quiz set related to the question: \\
\{quiz\_set\} \\

Please start by providing a step-by-step reasoning analysis of the quiz set, then return your evaluation as a JSON object in the following format: \\
'''json \\
\{ \\
"Educational Value": score, \\
"Diversity": score, \\
"Area Relevance": score, \\
"Difficulty Appropriateness": score, \\
"Comprehensiveness": score \\
\}'''
\end{tcolorbox}

Here is the prompt we use to do pairwise comparisons of quality of quiz set:
\begin{tcolorbox}[colframe=black, colback=gray!10, title=Prompt for Pairwise Comparison, breakable]

A student studying \{area\} at the \{level\} level has asked the following question: "\{question\}". You are given two quiz sets that aim to help the student better understand the question. Please choose the quiz set that best address this question. Please evaluate and compare the educational quality of these quiz sets based on the criteria listed below. For each criterion, select the quiz set that performs better by outputting 1 or 2. \\

1. Educational Value: Which quiz set offers greater learning potential? Which set will help students gain a deeper understanding of the topic? \\
2. Diversity: Which quiz set covers a broader range of topics? Does it explore a variety of concepts or focus narrowly on a single idea? \\
3. Area Relevance: Which quiz set is more aligned with the student's question and the key concepts they are studying? How well is it tailored to the specific subject area? \\
4. Difficulty Appropriateness: Which quiz set is better suited to the student's current educational level, neither too simple nor too advanced? \\
5. Comprehensiveness: Which quiz set provides greater depth and breadth? Which one is more thorough in addressing key concepts and details? \\

Here is the quiz set 1: \\
\{quiz\_set\_1\} \\

Here is the quiz set 2: \\
\{quiz\_set\_2\} \\

Please start by providing a step-by-step reasoning analysis of the quiz sets, then return your evaluation as a JSON object in the following format: \\
'''json \\
\{ \\
"Educational Value": choice, \\
"Diversity": choice, \\
"Area Relevance": choice, \\
"Difficulty Appropriateness": choice, \\
"Comprehensiveness": choice \\
\}'''
\end{tcolorbox}

\section{Additioanl Experimental Results with Gemini}
\label{apd:exp}

\begin{figure}[ht]
    \centering
    \includegraphics[width=\linewidth]{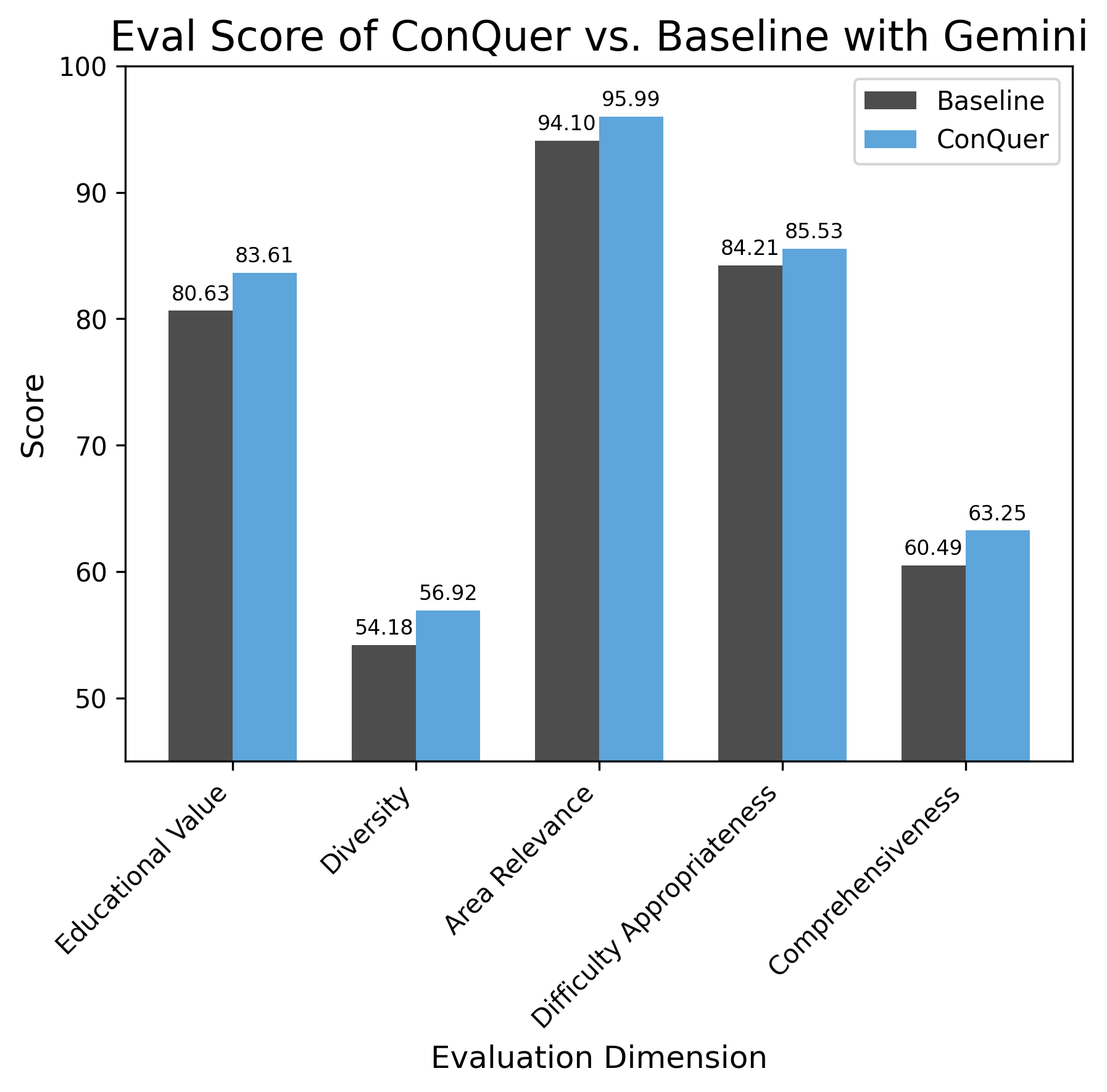}
    \caption{Evaluation score comparison between the baseline and ConQuer with Gemini-2.0-flash}
    \label{fig:apd1}
\end{figure}

\begin{figure}[ht]
    \centering
    \includegraphics[width=\linewidth]{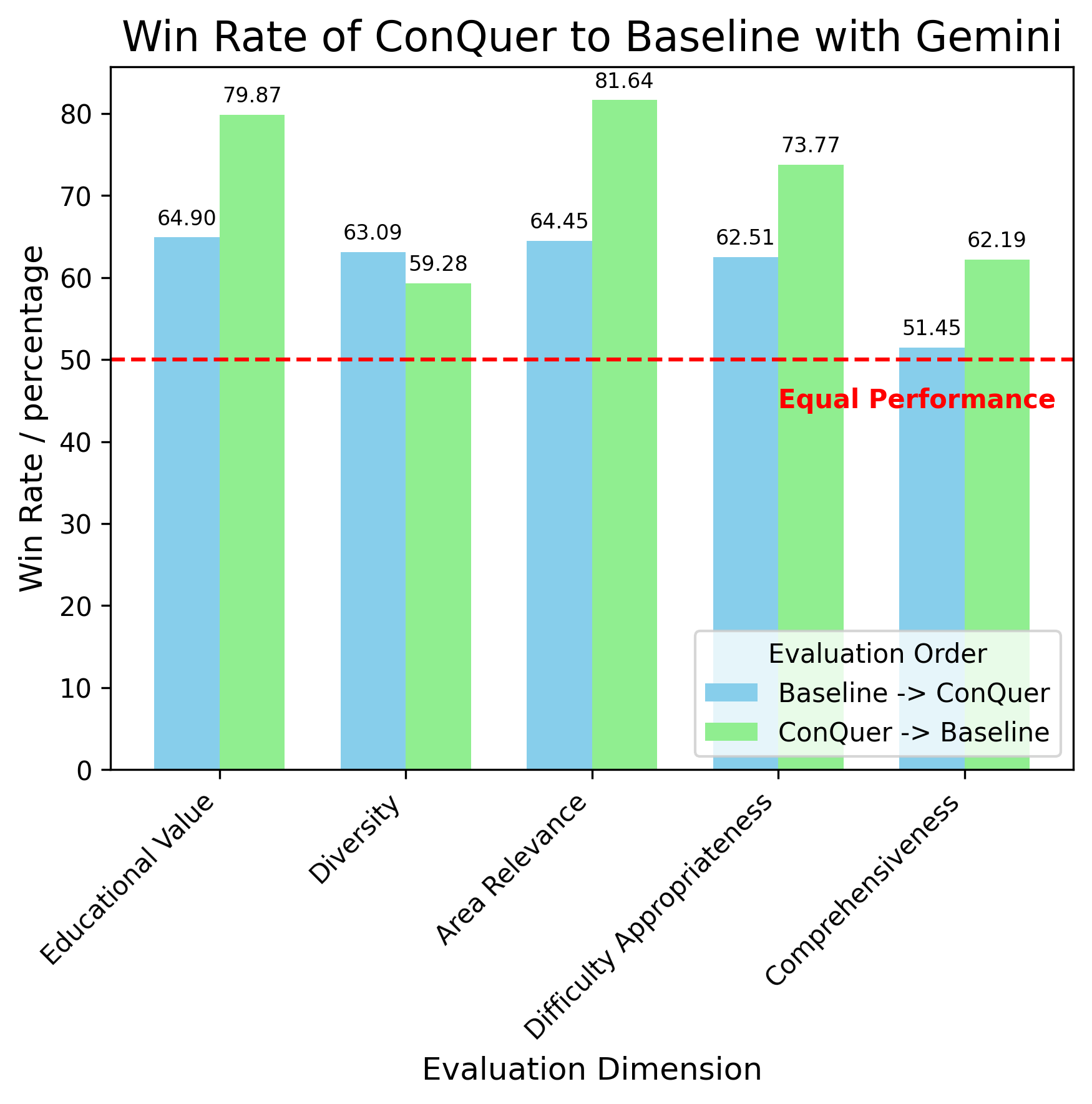}
    \caption{Win rate from pairwise comparison between the baseline and ConQuer with Gemini-2.0-flash}
    \label{fig:apd2}
\end{figure}

To further demonstrate the generalizability of ConQuer across different LLMs, we conducted experiments using Gemini-2.0-flash ~\cite{team2023gemini}. The corresponding results are presented in Figures ~\ref{fig:apd1} and ~\ref{fig:apd2}. On average, ConQuer achieved a 3.1\% improvement across five evaluation dimensions, with a win rate of 66.32\%. While this performance is slightly lower than that of GPT-4o-mini, it clearly demonstrates the effectiveness of the ConQuer framework in generating high-quality quizzes.

\section{Qualitative Analysis of Ablation Study}
\label{apd:qua}

In this section, we provide a qualitative analysis of the quizzes generated by ConQuer and compare them to quizzes generated with different modules removed, as shown in Table~\ref{tab:qual}. For clarity, we select representative quizzes from the quiz set and only present a subset of the results.

\begin{table*}[ht!]
\centering
\begin{tabular}{m{2cm}m{3cm}m{5cm}m{5cm}}

\toprule
\textbf{Ablation Module} & \textbf{Student Question} & \textbf{Quiz Generated without Ablation Module} & \textbf{Quiz Generated by ConQuer} \\
\midrule
Concept Extraction Module & What happens to a plant when it doesn't get enough sunlight or water? & Nothing about photosynthesis & Quiz: Which process do plants struggle with when they do not get enough sunlight? \newline
A. Photosynthesis \newline
B. Respiration \newline
C. Digestion \newline
D. Germination \\
\midrule
Wikipedia Knowledge Source & How does gravity affect the way objects move on Earth and in space? & Quiz: What force pulls objects toward each other on Earth? \newline
A. Gravity \newline
B. Magnetism \newline
C. Friction \newline
D. Electricity & Quiz: What determines the strength of Earth's gravitational field at a given location? \newline 
A. The mass of Earth and the distance from its center \newline
B. The temperature of the air \newline
C. The speed of sound in water \newline
D. The color of the sky \\
\midrule
Summarization Module & What role does childhood development play in shaping adult behavior and personality? & Quiz: Which of the following attachment styles was NOT identified by Mary Ainsworth in her strange situation experiment?  \newline
A. Independent \newline
B. Secure \newline
C. Ambivalent \newline
D. Avoidant & Quiz: What term is often used to describe personality in children? \newline
A. Temperament \newline
B. Mood \newline 
C. Character \newline
D. Disposition \\
\bottomrule
\end{tabular}
\caption{Ablation Study Result}
\label{tab:qual}
\end{table*}

Removing the concept extraction module significantly impacts the quiz's ability to capture the underlying concept behind the student's question. In the example related to plant growth, the quiz generated without this module fails to mention photosynthesis, which is essential for the student's understanding of the process and its importance for plants.

When the knowledge source is altered, the generated quiz becomes overly simplistic, essentially repeating basic concepts without depth. In contrast, the quiz generated by ConQuer, utilizing a more comprehensive knowledge base like Wikipedia, incorporates richer details, such as explaining how Earth's mass and distance influence gravity.

Finally, when the summarization module is removed, the resulting quiz deviates from the student's original question, likely due to the model's failure to focus on the key information. On the other hand, the quiz generated by ConQuer maintains a close alignment with the student's question, demonstrating its ability to stay on topic and provide relevant information.

\section{Analysis of Correlation of Evaluation Dimensions}
\label{apd:cor}
Given the range of evaluation dimensions employed in this study, it is essential to examine the relationships between them. To facilitate this, we present a heatmap illustrating the correlation between the scores of each evaluation dimension.

\begin{figure}
    \centering
    \includegraphics[width=\linewidth]{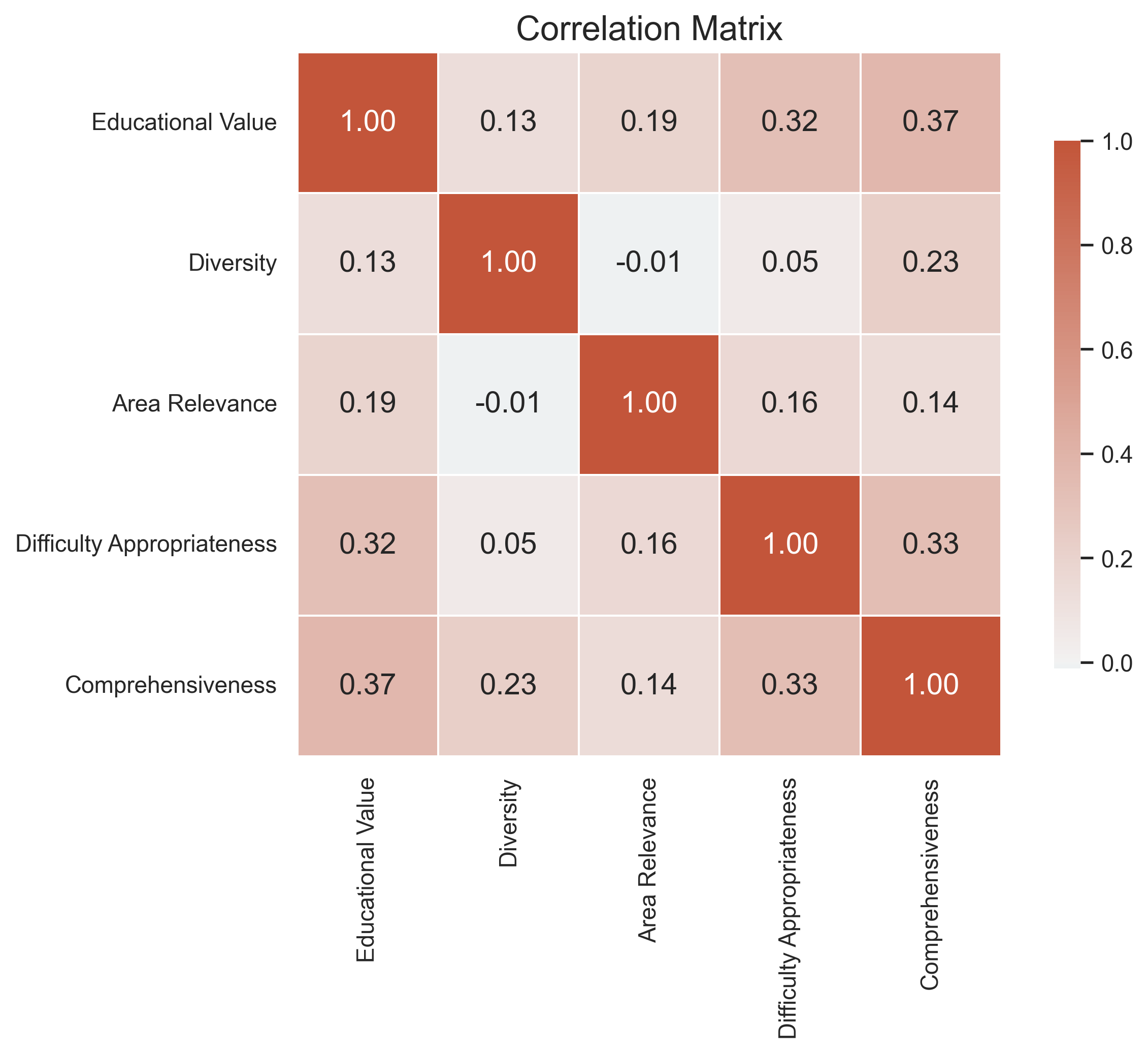}
    \caption{Correlation of scores in each evaluation dimension.}
    \label{fig:enter-label}
\end{figure}

The heatmap reveals several dimensions with strong positive correlations. For instance, Educational Value, Difficulty Appropriateness, and Comprehensiveness are closely related. These correlations can be explained by the fact that a more comprehensive quiz tends to cover a broader range of topics, thereby enhancing its educational value. Similarly, a difficulty level aligned with the student's abilities tends to improve both educational value and comprehension by appropriately challenging the learner.

On the other hand, some metrics exhibit little to no correlation. For example, Diversity and Area Relevance show near-zero or even negative correlations. This may occur because increasing the diversity of content often necessitates expanding the scope of topics, which could inadvertently reduce the focus on a specific subject area.

\end{document}